\documentclass{article} %
\usepackage{iclr2021_conference,times}
\iclrfinalcopy

\usepackage{xcolor}
\usepackage{hyperref}       %
\usepackage{url}
\hypersetup{
	colorlinks,
	linkcolor={red!50!black},
	citecolor={blue!50!black},
	urlcolor={blue!80!black}
}
\definecolor{orange}{HTML}{ff7f0e}
\definecolor{blue}{HTML}{1f77b4}

\usepackage{algorithm}
\usepackage[noend]{algpseudocode}
\usepackage{algorithmicx}

\usepackage{microtype}
\usepackage{lipsum}
\usepackage{graphicx}
\usepackage{subcaption}
\usepackage{sidecap}
\usepackage{caption}
\usepackage{booktabs} %
\usepackage{amsfonts}       %
\usepackage{nicefrac}       %
\usepackage{microtype}      %
\usepackage{comment}
\usepackage{enumitem}
\usepackage{wrapfig,tikz}
\usepackage{amsmath,longtable,fancyhdr}
\usepackage{bigstrut, tabularx, multirow, makecell, diagbox}
\usepackage[export]{adjustbox}
\usepackage{graphicx,float}
\usepackage{amssymb,amsthm, mathtools}
\usepackage{stackrel}
\usepackage{color}
\usepackage{colortbl}
\usepackage{theoremref}
\usepackage{cases}
\usepackage{stmaryrd}
\usepackage{mathabx}
\usepackage{dsfont}
\usepackage[capitalise]{cleveref}

\usepackage{pifont}

\newcommand{\mbb}[1]{\mathbb{#1}}
\newcolumntype{C}[1]{>{\centering\arraybackslash}m{#1}}
\newcolumntype{P}[1]{>{\centering\arraybackslash}p{#1}}

\newcommand{\mcal}{\mathcal}
\newcommand{\norm}[1]{\left\lVert#1\right\rVert}

\usepackage{mdframed}

\newtheorem{theorem}{Theorem}

\newtheorem{proposition}{Proposition}

\newcommand{\be}{\begin{equation}}
	\newcommand{\ee}{\end{equation}}
\definecolor{Gray}{gray}{0.85}
\definecolor{LightCyan}{rgb}{0.88,1,1}
\DeclareMathOperator*{\argmin}{arg\,min}
\DeclareMathOperator*{\argmax}{arg\,max}

\makeatletter
\usepackage{xspace} 
\def\@onedot{\ifx\@let@token.\else.\null\fi\xspace}
\DeclareRobustCommand\onedot{\futurelet\@let@token\@onedot}

\newcommand{\bfx}{\mathbf{x}}

\newcommand{\bfI}{\mathbf{I}}

\newcommand{\bfzero}{\mathbf{0}}

\newcommand{\bftheta}{{\boldsymbol{\theta}}}

\newcommand{\bfG}{\mathbf{G}}
\newcommand{\bfD}{\mathbf{D}}
\newcommand{\bfS}{\mathbf{S}}

\newcommand{\pd}{p_{\mathrm{data}}}

\definecolor{blue1}{RGB}{0,128,255}
\definecolor{blue3}{RGB}{0,0,128}
\definecolor{darkpastelgreen}{rgb}{0.01, 0.75, 0.24}
\definecolor{cerulean}{rgb}{0.0, 0.48, 0.65}

\definecolor{darkgreen}{rgb}{0,0.6,0}

\title{ Score Mismatching for Generative Modeling}

\author{Senmao Ye\\
	South China University of Technology\\
	\texttt{senmaoy@gmail.com}\\
	\And 
	Fei Liu\\
	South China University of Technology\\
	\texttt{feiliu@scut.edu.cn}\\
}

\begin{document}

\maketitle

\begin{abstract}
We propose a new score-based model with one-step sampling. Previously, score-based models were burdened with heavy computations due to iterative sampling. For substituting the iterative process, we train a standalone generator to compress all the time steps with the gradient backpropagated from the score network. In order to produce meaningful gradients for the generator, the score network is trained to simultaneously match the real data distribution and mismatch the fake data distribution. This model has the following advantages: 
 1) For sampling, it generates a fake image with only one step forward. 
 2) For training, it only needs  10 diffusion steps.
 3) Compared with consistency model, it is free of the ill-posed problem caused by consistency loss.
On the popular CIFAR-10 dataset, our model outperforms Consistency Model and Denoising Score Matching, which demonstrates the potential of the framework. We further provide more examples on the MINIST and LSUN datasets. The code is available on  \href{https://github.com/senmaoy/Score-Mismatching}{\textcolor{blue}{GitHub}}.
\end{abstract}

\section{Introduction}
 Recently, score-based models~\cite{ho2020denoising,DBLP:conf/nips/SongE19,DBLP:conf/iclr/0011SKKEP21,hyvarinen2005estimation}  achieved impressive generation performance across various tasks. Score-based models corrupt data samples with forward diffusion to get a set of noisy surrogates of the distribution gradient. Then, for sampling, score-based models predict the distribution gradients one after another. The iterative sampling process causes a heavy computational burden and blocks its application, especially in mobile applications. In order to accelerate score-based models, many researchers are interested in reducing the number of sample iterations~\cite{DBLP:journals/corr/abs-2303-01469,DBLP:conf/iclr/SongME21,DBLP:conf/nips/0011ZB0L022}. Among these models, Consistency model~\cite{DBLP:journals/corr/abs-2303-01469} first explores the potential of score-based models for one-step generation.  

 However, consistency model directly maps each diffusion step to the clean sample which caused an ill-posed problem. Because the random states in a Markov chain are not one-to-one matched.  Such principle also explains the phenomenon that predicting the expectation of each time step performs worse than predicting the noisy distribution score~\cite{ho2020denoising}.  Similarly, such a problem caused by biased   regression  is called the reconstruction bias in Cycle-GAN~\cite{DBLP:conf/iccv/ZhuPIE17}. In a word, our purpose is to develop a bias-free, one-step, and score-based generative model.

\begin{SCfigure}
	\centering
	\caption{ The illustration of the proposed score mismatching. SMM  only corrupts the data slightly with a few diffusion steps and samples with one-step forward.}
	\includegraphics[width=0.68\linewidth, trim=0 0 0 0, clip]{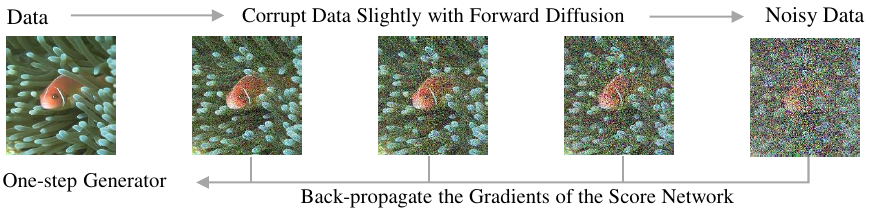}
	\label{fig:1}
\end{SCfigure}

In this paper, for substituting the iterative sampling process, a standalone generator is trained to compress all the time steps into one step. As depicted in Fig~\ref{fig:1}, our intuition is to supervise the generator with the the gradient of score network rather than channel activations from pre-trained models~\cite{DBLP:conf/nips/0011ZB0L022,DBLP:conf/iclr/SalimansH22}. However, there are two main challenges with this approach. First, the ordinary score network is only trained on true data with a domain gap compared to the learned data. Directly fitting the gradient of the score network leads to meaningless output. Second, removing the iterative sampling process leads to noise corruption. The score network is trained only on corrupted data samples, and clean data never appears during the training. Existing score-based models remove noise corruption with elaborate iterations. Hence, the gradient from the score network will force the generator to copy these noise corruptions.

To overcome the above two challenges, we propose a new score-based model called score mismatching (SMM). For bridging the domain gap between real and fake samples,  the score network is simultaneously trained to mismatch the fake data distribution. Hence, the gradient of the score network teaches the generator to be closer to the real data distribution and away from the fake data distribution. To eliminate the noise leaking problem, SMM removes the noise during training rather than in the testing phase. We carefully design a zero-mean noise injection pipeline. Then the generator is trained to output the same data sample with different noises, which makes the noises to cancel themselves.

SMM has several desirable properties. For sampling, its speed is much faster than previous score-based models because it needs only one step to infer a data sample. SMM is even faster than the Consistency Model because its generator is a deconvolution network rather than an U-net. For training, SMM spent less time than other score-based models because 10 diffusion steps are enough to get good performance. Thirdly, compared with Consistency Model, it is free of the ill-posed problem caused by consistency loss. At last, SMM constructs stronger adversarial training~\cite{goodfellow2014generative} by matching and mismatching the data distribution. The score network has to learn the data structure to corrupt and reconstruct an image like masked pre-training~\cite{DBLP:conf/cvpr/HeCXLDG22}, which provides more supervision than the adversarial loss of binary classification.
\section{Background}
\subsection{Denoising Score Matching}\label{sec:ncsn}
Score Matching\citep{hyvarinen2005estimation} estimates an unknown distribution by fitting the gradients of the log-density according from the observed data. To avoid computing the gradient of the log-density, Denoising Score Matching (DSM) \citep{DBLP:conf/nips/SongE19} proposes to substitute the distribution gradient with random Gaussian noise of
various magnitudes. Then the score network is trained to approximate such noisy surrogate of data distribution~\cite{vincent2011connection}. The score network $\bfS$, parameterized by $\theta$ and conditioned on the noise level $\sigma$, is tasked to minimize the following loss:
\begin{align}
\bftheta^*_{\bfS} &= \argmin_\bfS
   \sum_{i=1}^{N} \sigma_i^2  \mbb{E}_{\pd(\bfx)}\mbb{E}_{p_{\sigma_i}(\tilde{\bfx} \mid \bfx)}\big[\norm{ \bfS(\tilde{\bfx}, \sigma_i) - \nabla_{\tilde{\bfx}} \log p_{\sigma_i}(\tilde{\bfx} \mid \bfx)}_2^2 \big],\label{eqn:ncsn_obj}
\end{align}
where  $p(\boldsymbol{x})$ the training data distribution, and $p(\sigma)$ the uniform distribution over a set $\{\sigma_i\}$ corresponding to different levels of noise and $\tilde{\bfx}$ is a data sample corrupted by Gaussian noise $p_{\sigma}(\tilde{\bfx} \mid \bfx) \coloneqq \mcal{N}(\tilde{\bfx}; \bfx, \sigma^2 \bfI)$.

\subsection{Generative Adversarial Network}\label{sec:gan}
A generative adversarial network consists of a generator and a discriminator. The training objective for the discriminator is a binary classification loss, which takes the synthesized image as negative samples and real images as positive samples. The corresponding training objective of the generator is cheating the discriminator that the fake samples are real:
\begin{align}
	\bftheta^*_\bfD &= \argmax_\bfD {\rm log} \bfD(x) + \ {\rm log}(1-\bfD(\bfG(z)))\\
	\bftheta^*_\bfG &= \argmin_\bfG  \ {\rm log}(1-\bfD(\bfG(z))),
\end{align}

where z is the random noise sampled from the Gaussian distribution. 

\section{Score Mismatching}

\subsection{Generative Modeling with Score Mismatching}\label{sec:sde_examples}
In this section, we aim to build a score-based model with only one step forward for sampling. To achieve this goal, we compress all the time steps with a standalone generator. Hence, our SMM is composed of a score network S and a generator G. Specifically, $\bfS$ is trained to simultaneously match the score of true data distribution and mismatch the fake data distribution; G is trained to cheat S to match the score of fake data distribution. The time step $t$ is fed into $\bfS$ every time, but we omit it for convenience.  The training objective of score network S and generator G is:

\begin{align}
	\bftheta^*_\bfS &= \argmin_\bfS \norm{ \bfS(\bfx+\epsilon_1\sigma_t)-\epsilon_1}_2^2 + \argmin_\bfS \norm{ \bfS(\bfG(z_1)+\epsilon_2\sigma_t)-\epsilon_3}_2^2\\
		\bftheta^*_\bfG &= \argmin_\bfG \norm{ \bfS(\bfG(z_2)+\epsilon_4\sigma_t)-\epsilon_4}_2^2\label{eq:G},
\end{align}
where $z_1,z_2$ is a random vector from standard Gaussian distribution. $\bfG(z)$ draws a data sample from the learned distribution of the generator. $\epsilon_1,\epsilon_2,\epsilon_3,\epsilon_4$ are the same size as x and independently sampled from standard Gaussian distribution.  Notably, $z_1, z_2$ must be sampled independently, or the training of S and G will immediately conflict with each other. Following DDPM~\cite{ho2020denoising}, $\sigma_t$ is the noise variance at each diffusion step:
\begin{align}
\sigma_t = \sqrt{1-\bar{\alpha}_{t}}, \quad\quad\quad
\bar{\alpha}_{t}=\prod_{i=1}^{T} 1-\beta_i,
\end{align}
where $\beta_t$ is set increasing from $\beta_0 = 1e-4$ to $\beta_{10}=0.02$. 
The detailed training algorithm is shown below:

\begin{minipage}[t]{\linewidth}
	\begin{algorithm}[H]
		\caption{Minibatch stochastic gradient descent training of score mismatching.}
		\label{alg:corrector_ve}
		\begin{algorithmic}[1]
			\Require{ training iteration $N$, diffusion steps $T$, noise variance $\{\sigma_i\}_{i=1}^T$.}
                \For{$j=1$ to $n$}
			\State{$z_1,z_2,\epsilon_1,\epsilon_2,\epsilon_3,\epsilon_4 \sim \mcal{N}(\bfzero,  \bfI)$ and $t\sim (0,T)$ \quad Generate random variables independently}
			\State{$\bftheta^*_\bfS = \argmin_\bfS \norm{ \bfS(\bfx+\epsilon_1\sigma_t)-\epsilon_1}_2^2$ \quad \quad\quad\quad Match true data distribution}		
			\State{$\bftheta^*_\bfS = \argmin_\bfS \norm{ \bfS(\bfG(z_1)+\epsilon_2\sigma_t)-\epsilon_3}_2^2$ \quad \quad Mismatch fake data distribution}
			\State{$\bftheta^*_\bfG = \argmin_\bfG \norm{ \bfS(\bfG(z_2)+\epsilon_4\sigma_t)-\epsilon_4}_2^2$ \quad \; The generator fit the score network}
				\EndFor
			\Return{$\bfS, \bfG$}
		\end{algorithmic}
	\end{algorithm}
\end{minipage}

\subsection{The Global Optimal of Score Mismatching}
In this section, we show that SMM is guaranteed to converge to the true data distribution if the stochastic training is optimal. To achieve this goal, we  first consider the optimal score network S for any given generator G.
\begin{proposition}[]\label{thm:score}The global optimal of score network is:
	\begin{align}
		\frac{p_{data}\epsilon_1+p_g\epsilon_3}{p_{data}+p_g}
	\end{align}
\begin{proof}
	The training criterion for the score network S, given any generator G, is to maximize the
	$\ell^2$ loss:	
	\begin{align}
		l_{\bfS} &= \int_x \int_{\epsilon_1} p_{data}(x) \norm{ \bfS(\bfx+\epsilon_1\sigma_t)-\epsilon_1}_2^2 + \int_z \int_{\epsilon_2,\epsilon_3} p_z(z) \norm{ \bfS(\bfG(z_1)+\epsilon_2\sigma_t)-\epsilon_3}_2^2\\
		&= \int_x p_{data}(x) \int_{\epsilon_1}\norm{ \bfS(\bfx+\epsilon_1\sigma_t)-\epsilon_1}_2^2 + p_g(x) \int_{\epsilon_2,\epsilon_3} \norm{ \bfS(x+\epsilon_2\sigma_t)-\epsilon_3}_2^2\\
		&= \int_{x,\epsilon_1,\epsilon_3} p_{data}(x)\norm{ \bfS(\bfx+\epsilon_1\sigma_t)-\epsilon_1}_2^2 + p_g(x) \norm{ \bfS(x+\epsilon_1\sigma_t)-\epsilon_3}_2^2\label{thm:quadratic}
	\end{align}
In Equation~\ref{thm:quadratic}, $p_{data}(x)\norm{ \bfS(\bfx+\epsilon_1\sigma_t)-\epsilon_1}_2^2 + p_g(x) \norm{ \bfS(x+\epsilon_1\sigma_t)-\epsilon_3}_2^2$ is  a quadratic function of $\bfS(\bfx+\epsilon_1\sigma_t)$, hence it  reaches its minimum when $\bfS(\bfx+\epsilon_1\sigma_t) = \frac{p_{data}\epsilon_1+p_g\epsilon_3}{p_{data}+p_g}$.
\end{proof}
\end{proposition}

 The score-based models are trained on corrupted data and existing works remove the noise corruption with elaborate iterations. Hence, without the iterative sampling procedure, SMM is exposed to the pollution of heavy noise corruption. To solve this problem, we introduce the following proposition:
\begin{proposition}[]\label{thm:noise}The generator $\bfG$  is free of noise corruption if the noise corruption is zero-mean.

When $\bfG$  is trained to output $x+\epsilon_1\sigma_t$ with $\ell^2$ loss, it eventually outputs $x$.
$\ell^2$ loss is well-known for its characteristics of smoothness in image denoising~\cite{DBLP:conf/icml/LehtinenMHLKAA18}.	In the training of SMM, such smoothness can also eliminate the noise.
 Specifically, when the score network $\bfS$ receives a sample of $x+n_1$, it propagates $\partial \ell^2|S(x+n_1)-n_1|$. When $\bfS$ receives  $x+n_2$, it propagates $\partial \ell^2|S(x+n_2)-n_2|$. In the entire training procedure, the generator will be forced to produce $x+n_1+n_2+...+n_i= x +E(n) = x$.
	
	\end{proposition}

With the above two propositions, we are able to prove that the generator implicitly learns the true data distribution.
\begin{theorem}[]\label{thm:optimal} $p_g = p_{data}$ is the global minimum of the training criterion of G. At that point, $l_{\bfG}$ achieves the value $1$.

\begin{proof}	
According to the Jensen Inequality and Proposition~\ref{thm:score}, Equation~\ref{eq:G} has the following lower bound:
	\begin{align}
	l_{\bfG}	&=\int_{x,\epsilon_3,\epsilon_4}  p_g\norm{ \frac{p_{data}\epsilon_4+p_g\epsilon_3}{p_{data}+p_g}-\epsilon_4}_2^2\\
		&= \int_{x,\epsilon_3,\epsilon_4}   p_g\norm{ \frac{p_{g}\epsilon_3-p_{g}\epsilon_4}{p_{data}+p_g}}_2^2\\
		&\geq(\int_{x,\epsilon_3,\epsilon_4}  p_g \frac{p_{g}}{p_{data}+p_g} \times  (\epsilon_3-\epsilon_4) )^2\\
		&\geq (\frac{\int_x p_g p_g  }{\int_x p_g p_g + \int_x p_g p_{data}})^2 \times \int_{\epsilon_3,\epsilon_4} (\epsilon_3-\epsilon_4)^2\\
  &\geq (\frac{1 }{1 + \int_x p_g \frac{p_{data}}{p_g}})^2 \times \int_{\epsilon_3,\epsilon_4} (\epsilon_3-\epsilon_4)^2 ,
	\end{align}
where $\epsilon_3$ and $\epsilon_4$ are from Gaussian distribution hence the expectation of $(\epsilon_3-\epsilon_4)^2$ is a constant of $4$.  When $p_g= p_{data}$, $l_{\bfG} =1$ which is the lower bound of $l_{\bfG}$. 
\end{proof}
\end{theorem}
Considering that the observed data distribution is polluted by noises, the generator is trained to output $p_{data}+\epsilon\sigma_t$. However, considering Proposition 2, the generator finally outputs the clean data distribution $p_{data}$.

\subsection{Alternative Variants of Score Mismatching}
 As the stochastic training is not optimal, SMM can not fully remove the noise corruptions. To circumvent this problem, we introduce another variant of score mismatching. This variant of SMM directly trains the score network on clean data so that the generator is trained to learn the clean data distribution. It is similar to the first diffusion model~\cite{sohl2015deep} that predicts noisy images rather than the distribution gradient:
\begin{align}
	\bftheta^*_\bfS &= \argmin_\bfS \norm{ \bfS(\bfx,\epsilon_1\sigma_t)-(\bfx+\epsilon_1\sigma_t)}_2^2 + \argmin_\bfS \norm{ \bfS(\bfG(z_1),\epsilon_2\sigma_t)-(\bfG(z_1)+\epsilon_3\sigma_t)}_2^2\\
	\bftheta^*_\bfG &= \argmin_\bfG \norm{ \bfS(\bfG(z_2),\epsilon_4\sigma_t)-(\bfG(z_2)+\epsilon_4\sigma_t)}_2^2
\end{align}
This variant is more robust to heavy noise corruption but the final performance is poorer than the original SMM. We conjugate that $\bfx$ and $\bfx+\epsilon\sigma_t$ are not one-to-one matched, and directly minimizing the $\ell_2$ loss between them leads to an ill-posed problem~\cite{bakushinsky2012ill,DBLP:conf/cvpr/JoOVK21}. That's why the simple loss proposed by DDPM~\cite{ho2020denoising} leads to significant improvement in image quality.

A better solution is to inject a clean image and a noisy image simultaneously into the score network, which avoids direct optimization toward ill-posed output:
\begin{align}
	\bftheta^*_\bfS &= \argmin_\bfS \norm{ \bfS(\bfx,\bfx+\epsilon_1\sigma_t)-\epsilon_1}_2^2 + \argmin_\bfS \norm{ \bfS(\bfG(z_1),\bfG(z_1)+\epsilon_2\sigma_t)-\epsilon_3}_2^2\\
	\bftheta^*_\bfG &= \argmin_\bfG \norm{ \bfS(\bfG(z_2),\bfG(z_2)+\epsilon_4\sigma_t)-\epsilon_4}_2^2
\end{align}
However, this variant also works worse than the original SMM. We introduce these variants here to inspire other possibilities for improving the noise leaking problem. For convenience, we denote these two variants as SMM variant 2 and SMM variant 3~\label{variant}.
\section{Experiments}
\paragraph{Dataset} We conduct qualitative and quantitative experiments to evaluate the performance of the proposed SMM on three commonly used image datasets, i.e. CIFAR-10, LSUN, and MINIST. All the images are resized to the resolution of $32\times32$ and normalized to the range of $(-1,1)$. CIFAR-10 contains 50,000 images from 10 classes of natural images. For LSUN, we use its two sub-classes of bedrooms with 400,000 and cats of 240,000 images respectively. The MINIST dataset consists of 70,000 images of hand-written digital numbers.

\paragraph{Hyper Parameters}

For training the networks, we use Adam optimizer with base learning rate of 0.0025 and batch size of 32. The training of CIFAR-10 takes around 5 days on a single GTX 3090. The training process is stopped when the FID score does not descend on training data. For corrupting the images, we set $\beta_t$ to constants increasing linearly from $\beta_1 = 10^{-4}$ to $\beta_{10}=0.02$.
\subsection{Image Generation with SMM}
\begin{figure*}[h]

	\begin{subfigure}[t]{0.33\linewidth}
		\centering
		\captionsetup{justification=centering}
		\includegraphics[width=1.78in]{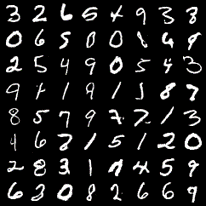}
		\caption{\ }

	\end{subfigure}%
	\begin{subfigure}[t]{0.33\linewidth}
		\centering
		\captionsetup{justification=centering}
		\includegraphics[width=1.78in]{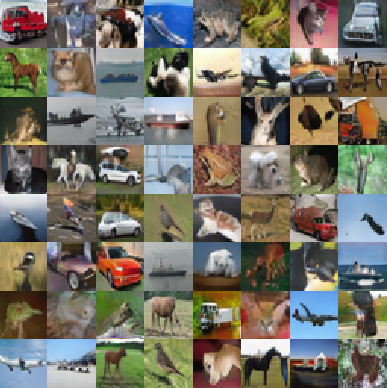}
		\caption{\ }

  	\end{subfigure}%
  	\begin{subfigure}[t]{0.33\linewidth}
		\centering
		\captionsetup{justification=centering}
		\includegraphics[width=1.78in]{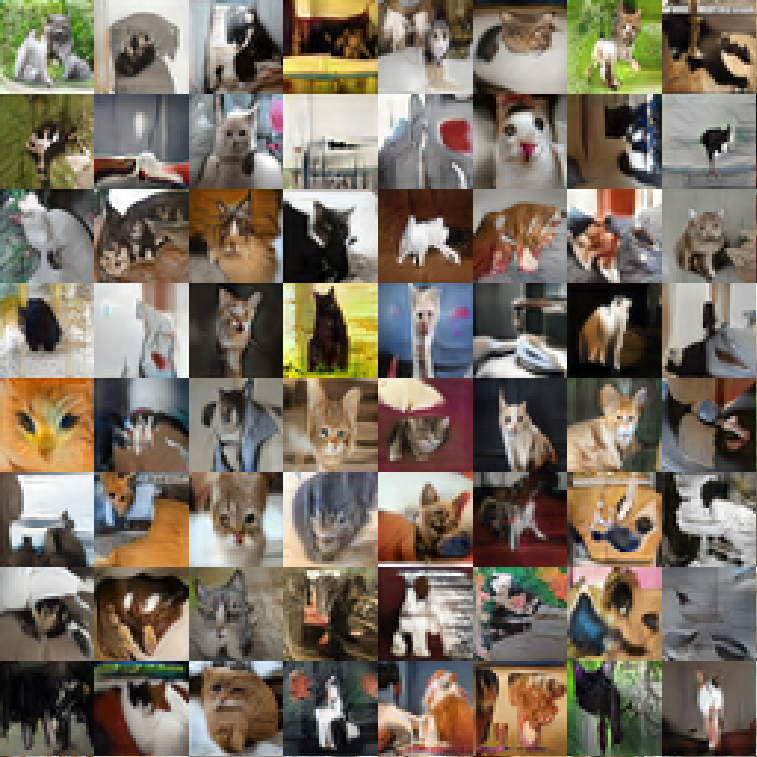}
		\caption{\  }

	\end{subfigure}%

	\vspace{-4pt}
	\caption{\label{Fig:sample}Qualitative results on different datasets. (a) Results from the MINIST dataset (b) Results from the CIFAR-10 dataset (c) Results from the LSUN-bedroom dataset.}
\end{figure*}

In Fig~\ref{Fig:sample}, we show uncurated samples of SMM on the  MNIST, LSUN, and CIFAR-10 datasets. SMM successfully synthesizes various categories of images which shows its generalization ability. As shown by the samples, our generated images have higher or comparable quality to those from previous score-base models and GANs. What's more, there is nearly no corrupted sample, which demonstrates the stability of the proposed SMM.

For quantitative evaluation, we report Inception Score (IS) and Fr\'{e}chet Inception Distance (FID) on CIFAR-10 in Table~\ref{tab_sota}. As a one-step and score-based model, we achieve the FID score of 8.1, which is even better than the multi-step score-based model~\cite{DBLP:conf/nips/SongE19} and GAN model~\cite{brock2018large}. Noticeably, our quantitative results outperform Consistency Model which is the best one-step and score-based model. Our FID score on CIFAR-10 is also comparable to top existing models, such as StyleGAN-v2~\cite{DBLP:conf/cvpr/KarrasLAHLA20}.

	\begin{table}
		\centering
		\caption{Quantitative comparison with state-of-the-art generative models on the CIFAR-10 dataset .}
		\label{tab_sota}
		\tabcolsep=0.25cm
		\begin{tabular}{ccccc}
			\hline
			Model & FID & IS & Score-based& One-step  \\
			\hline
			Generative Flow~\cite{DBLP:conf/nips/GrcicGS21}  & 34.90 & -  & \ding{55}  & \checkmark \\
			StyleGAN-v2~\cite{DBLP:conf/cvpr/KarrasLAHLA20}  & \textbf{5.02} & \textbf{10.17} & \ding{55}& \checkmark  \\
			BigGAN~\cite{brock2018large}  & 14.73 & 9.22 & \ding{55}& \checkmark  \\
			\hline
			Denoising Score Matching~\cite{DBLP:conf/nips/SongE19} & 25.32  & 8.87  & \checkmark & \ding{55} \\
			DDPM~\cite{ho2020denoising}& 3.17  & 9.46  & \checkmark & \ding{55} \\
              NCSPP~\cite{song2019generative}& \textbf{2.20}  & \textbf{9.89}  & \checkmark & \ding{55} \\
			\hline
			Consistency Model~\cite{DBLP:journals/corr/abs-2303-01469}  & 8.70 & 8.49   & \checkmark & \checkmark  \\
			Score Mismatching (Ours)  & \textbf{8.10} &\textbf{ 9.11}   & \checkmark & \checkmark  \\
			\hline
		\end{tabular}
	\end{table}

\subsection{The Influence of Different Noise Corruptions}
\paragraph{Noises}In this section, we explore the influence of different noise corruptions. In addition to the zero-mean noise used in SMM, we further explored two other noises, i.e. nonzero-mean noise, and spatial diffusion. Zero-mean noise $x+\epsilon\sigma_t$ is the most widely used corruption among score-based models. Nonzero-mean noise $\alpha_t(x+\epsilon\sigma_t)$~\cite{ho2020denoising} adding a declining scaling term to the image pixels.

The above two corruptions assume that image pixels are irrelevant to each other. However, the pixels will influence each other in the real diffusion process. We also tried to model such spatial diffusion within image pixels. In nature, the gas molecules make irregular thermal movements. Diffusion is the phenomenon of migration of substances due to the thermal movement of particles (molecules, atoms, etc.). The larger the concentration difference, the faster the diffusion. Hence,  unevenness in the spatial diffusion will cause a flow in the spatial dimension.  We simply assume that the speed of flow is proportional to the difference in corruption degree:
\begin{align}
    \Delta c = \sum_{\hat{c}} (\hat{c}-c) \times ratio ,
\end{align}
where $c\sim(0,1)$ is the corruption degree of each pixel. $\hat{c}$ is the corruption degree at neighboring pixels of $c$. The initial values of $c$  are $0$ except for the image center which is initialized as $1$. Then we update the corruption degree at every diffusion step. 

\paragraph{Results}The ablation studies of noise corruption are summarized in 
 Table~\ref{tab:noise}. Both non-zero mean Gaussian noise and spatial diffusion lead to worse performance because the noise corruptions are retained in the synthesized images.  In addition to noises, we also tried to flip the images horizontally which improves the FID score obviously.

\begin{table}[h]
	\centering
	\caption{The influence of different noise corruptions on CIFAR-10. Non-zero means Gaussian noise with non-zero mean. Spatial means Gaussian noise relevant to spatial location. Duplicate means duplicating each fake sample once. Flip means flipping the images horizontally.}
	\label{tab:noise}
	\tabcolsep=0.25cm
	\begin{tabular}{ccccccc}
		\hline
		ID & Non-zero &Spatial &Duplicate  &Flip&FID &IS  \\
		\hline
   		0 & - & - & -& -&14.45&8.29 \\

		1 & \checkmark & - &- & -& 24.78& 6.84\\
		
		2 & - & \checkmark &\- & -&65.30&4.56 \\
  		3 & - & - & \checkmark& -&12.36&8.50 \\
  		4 & - & - & -& \checkmark&13.43&8.36 \\
		5 & - & - &\checkmark & \checkmark&\textbf{10.11}&\textbf{8.97} \\\hline
	\end{tabular}
\end{table}

\begin{figure*}[h]

	\begin{subfigure}[t]{0.5\linewidth}
		\centering
		\captionsetup{justification=centering}
		\includegraphics[width=2.55in]{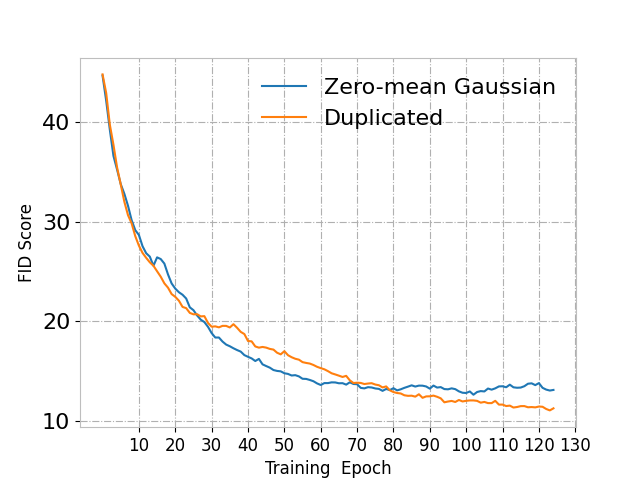}
		\caption{Enhance the generator by duplicating fake images}
		\label{fig:side:a}
	\end{subfigure}%
	\begin{subfigure}[t]{0.5\linewidth}
		\centering
		\captionsetup{justification=centering}
		\includegraphics[width=2.55in]{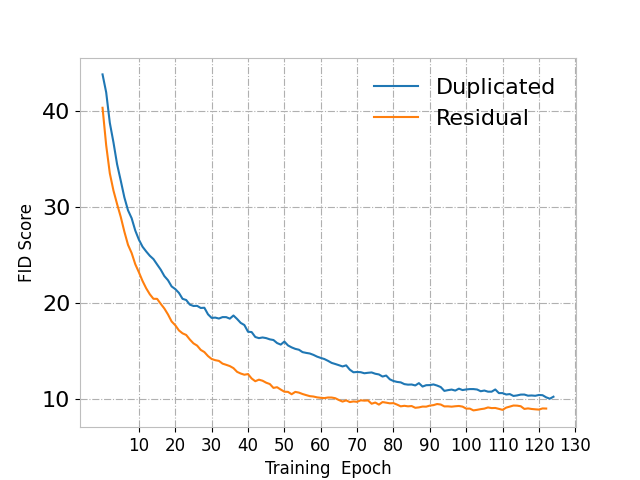}
		\caption{Adapt the score network with residual connections. }
		\label{fig:side:b}
	\end{subfigure}%

	\vspace{-4pt}
	\caption{ 	\label{Fig:curve} Visualization of the training curves of different ablation studies on CIFAR-10. FID score is used to quantitatively evaluate the model performance.
	}
\end{figure*}

\paragraph{Removing Noises Explicitly}
SMM removes the noise corruption according to Proposition 2, which produces a clean image by averaging over a bunch of noisy images. However, these noisy images are distributed in different batches, which makes the average operation 
 implicitly and inefficiently. To make this operation efficient, we train SMM with one more noisy image duplicated from each fake image. Then the training objectives of $\bfS$ and $\bfG$ become:
\begin{align}
\bftheta^*_\bfS &= \argmin_\bfS \norm{ \bfS(\bfx+\epsilon_1\sigma_t)-\epsilon_1}_2^2 + \argmin_\bfS \norm{ \bfS(\bfG(z_1)+\epsilon_2\sigma_t)-\epsilon_3}_2^2\\
	\bftheta^*_\bfG &= \argmin_\bfG \norm{ \bfS(\bfG(z_2)+\epsilon_4\sigma_t)-\epsilon_4}_2^2 + \norm{ \bfS(\bfG(z_2)+\epsilon_5\sigma_t)-\epsilon_5}_2^2, \label{eq:double}
\end{align}
where $e_5$ is sampled from the same distribution as $e_4$. Then the same fake image from G receives two different gradients caused by different noise corruptions. Although  Equation~\ref{eq:double} uses one more duplicated sample for training, it obviously improves the convergence speed and the FID score, as shown in Fig~\ref{fig:side:a}.

\subsection{Setting the Diffusion Steps}
	\begin{table}
	\centering
	\caption{The variation of  FID /IS score according to different diffusion steps on CIFAR-10.  }
	\label{tab_demo}
	\tabcolsep=0.25cm
	\begin{tabular}{ccccccc}
		\hline
		Diffusion Step & 0 &5 & 10 &100&500 &1000  \\
		\hline
		SMM & -/- & - /-  & \textbf{8.13/9.02} & 19.23/6.87&-/-&-/- \\

		Variant 2& -/- & -/- & 18.90/7.34 &\textbf{ 17.87/7.45}&24.56/6.90&26.89/6.75 \\
		
		Variant 3 & -/- & - /-  &\textbf{ 15.67/7.32} & 35.45/4.69&57.56/4.56&60.87/4.34 \\
		\hline
	\end{tabular}\label{tab:step}
\end{table}

The diffusion step is an important hyper-parameter for SMM. In Table~\ref{tab:step}, we show the model performance with different diffusion steps. More diffusion steps mean heavier noise corruptions. Specifically, $\beta_t$ increases linearly from $\beta_1 =10^{-4}$ to $\beta_T =0.02$.  We observe that: 1) Too much diffusion step leads to bad performance of SMM. Too heavy noise pollution even makes generators fail to recover a clean image.
2) The SMM variants 2 and 3  in section~\ref{variant} work well on heavy noises but perform worse than the original SMM. It seems that $\bfx$ and $\bfx+\epsilon\sigma_t$ are not one-to-one matched and directly minimizing such $\ell_2$ loss leads to an ill-posed problem. Maybe that's why the simple loss proposed by DDPM~\cite{ho2020denoising} leads to significant improvement in image quality. 
3)Too few diffusion steps also make the model not work, because the noises are almost nonexistent.

To sum up, different from diffusion models and score matching, SMM needs much fewer diffusion steps. Because it is naturally good at reducing the redundancy among diffusion steps.  It learns the image structure from matching and mismatching distribution scores, which is similar to Masked Autoencoders (MAE)~\cite{DBLP:conf/cvpr/HeCXLDG22}. 
\subsection{Ablation Studies on the Network Architecture}

	\begin{table}[!h]
	\centering
	\caption{Ablation studies on the Architecture of the score network on the CIFAR-10 dataset.}
	\label{tab_arc}
	\tabcolsep=0.15cm
	\begin{tabular}{cccccccc}
		\hline
		Model & NF=8& NF=16& NF=32&NF=64& Residual&RAT   & U-net \\
		\hline
		FID/IS  & 14.63/8/65 & 10.11/8.97  & 16.74/8.15 & -/- & 9.15/9.07 &\textbf{8.10/9.11}& 18.01/8.23  \\
		\hline

	\end{tabular}
\end{table}
SMM is composed of two networks: the generator and the score network. For the generator, we use the network architecture of Stylegan-v2~\cite{Karras2019stylegan2} in all the experiments. For the score network, we use the U-net from NCSPP~\cite{DBLP:conf/iclr/0011SKKEP21}. To study the influence of different capacities of the score network, we present results with a base dimension of NF=8, NF=16, NF=32, and NF=64 in Table~\ref{tab_arc}. It's obvious that a suitable base dimension is crucial for good performance and too large base dimensions corrupt the training.  - means the result is not available because the training collapses. To study the influence of different U-nets, we also tried a vanilla U-net implemented by us. In Table~\ref{tab_arc}, our implementation performs much worse than NCSPP which reveals that the network architecture is important to the performance. However, the potential of our model is not fully revealed due to limited computational resources. 

We also improve SMM with recurrent affine transformations~\cite{ye2023recurrent} and residual connections~\cite{he2016deep}. In Table~\ref{tab_arc}, Residual means NCSPP with NF=16 and residual connections. According to our experiment, four layers of residual connections between each resolution achieve the best results. When the number of residual connections is bigger than 8, the training is unstable. From Fig~\ref{fig:side:b}, it's obvious that residual connection accelerates the training process and leads to better results.
RAT means the aforementioned Residual variant with RAT. The RAT variant adds a RAT layer after each convolution operation in the generator.

%
%
%
%
%
%

\section{Related Work}
In this section, we introduce the relationship between SMM and existing generative models. 
SMM is different from previous score-based models in the following aspects: 1) For sampling, SMM only needs one step forward. 2) For training, SMM only needs a few diffusion steps. 3) For denoising, SMM makes the noises cancel with each other rather than inverse diffusion.

Noticeably, SMM reveals a conjugation relationship between adversarial and non-adversarial Models. 
SMM can be viewed as the hybrid model of score matching and GAN. It is possible that there are more conjugate models for existing models.  The work of~\cite{DBLP:journals/cagd/LeiSCYG19} also shows that adversarial training is not necessary. Compared with GAN, SMM reconstructs and destroys the distribution rather than playing a  classification game. There are also works~\cite{DBLP:conf/iclr/Jolicoeur-Martineau21,DBLP:conf/icml/KimKKKM23} that directly use GAN to refine score-based models by predicting better scores. However, these works still adopt the iterative sampling strategy.

The same as SMM, Consistency Model also belongs to score-based model and inference with one-Cstep forward, but SMM is free of its ill-posed problem. 
Consistency Model directly maps different time steps to the clean image but different time steps are not naturally one-to-one matched. Such principle also explains the phenomenon that predicting the expectation of each time step performs worse than predicting the noisy distribution score~\cite{ho2020denoising}.
Similar to Consistency Model, SMM also has the potential to get better performance with the supervision from pre-trained score-based models. Specifically, if a pre-trained model could provide accurate gradients of the target distribution, we can train SMM on clean images directly. However, current score-based models are usually built upon denoising score matching and output the noisy surrogate of the distribution gradient.
\section{Conclusion and Future work}
In this paper, we present a new score-based model with one-step sampling.
For substituting the iterative process, we train a standalone generator to compress all the time steps with the gradient back-propagated from the score network. To remove the diffusion noises, we make the zero-mean noises cancel each other. On the popular CIFAR-10 dataset, our model outperforms Consistency Model and Denoising Score Matching, which demonstrates the potential of the framework. We further elaborate on the conjugation relationship between adversarial and non-adversarial models.

In the future, there is much room to improve SMM. Specifically, we are interested in score estimation method without noises and adversarial conjugation models of other non-adversarial models.

\bibliography{SMM}
\bibliographystyle{iclr2021_conference}
\appendix

\end{document}